%% file: main.tex
\documentclass{article}
\usepackage{microtype}
\usepackage{graphicx}
\usepackage{booktabs}
\usepackage{hyperref}

\usepackage[preprint]{icml2026}
\usepackage{iftex}
\ifXeTeX\usepackage{fontspec}\fi
\usepackage{amsmath,amssymb}
\usepackage[capitalize,noabbrev]{cleveref}
\usepackage{listings}
\icmltitlerunning{Same Papers, Different Verdicts}

\begin{document}
\twocolumn[
\icmltitle{Same Papers, Different Verdicts:\\Tracing Version Drift in LLM Peer Review}
\icmlsetsymbol{equal}{*}
\begin{icmlauthorlist}
\icmlauthor{Jizhou Guo}{ind}
\end{icmlauthorlist}
\icmlaffiliation{ind}{Independent Researcher}
\icmlcorrespondingauthor{Jizhou Guo}{mitsuha2021b@gmail.com}
\icmlkeywords{LLM-as-a-judge, peer review, evaluation, model versions, contamination}
\vskip 0.3in
]
\printAffiliationsAndNotice{}
\hypersetup{pdftitle={Same Papers, Different Verdicts: Tracing Version Drift in LLM Peer Review},pdfsubject={Preprint}}

\begin{abstract}
\input{abstract}
\end{abstract}

\input{sec_intro}
\input{sec_setup}
\input{sec_results}

\input{sec_locate}
\input{sec_prereg}
\input{sec_related}
\input{sec_discussion}
\input{sec_statements}

\bibliography{refs}
\bibliographystyle{icml2026}

\input{sec_appendix}
\end{document}

%% file: abstract.tex
Venues and benchmarks read LLM review scores as measurements of a paper, while providers replace their models within months. We audit twenty frontier models from OpenAI, Google and Anthropic on ICLR 2027 submissions while they were under review, with tests registered before the decisions. Between an earlier and a later version of Claude Opus and of Gemini, the share of papers placed on the accept side falls by 52 and 53 percentage points on 183 and 60 randomly drawn submissions, and by 43 to 79 points in six further paired comparisons, with no paper moving up in any of them. Stating a 30\% acceptance rate in the prompt leaves a gap of 79 points between the two Opus versions. When score-stripped critiques are swapped between versions, the later critiques alone lower the accept share by 35 points for Opus and 51 for Gemini; for Opus the drop is 35 points under a third-party scorer and 24 after a rewrite that keeps each technical point and removes evaluative wording. On 57 papers, the percentiles of Opus 4.6 and 5.5 correlate at 0.73 while their accept shares differ by 44 points. On 99 to 119 submissions, the self-estimated percentiles of three models correlate with two human reviewers more strongly than a third reviewer does (point estimates 0.17 to 0.21, 95\% intervals between $-0.04$ and $+0.42$). All twenty models give a 2 or a 3 in 473 of their 475 reviews, against 60 of 72 human reviews. Retrospective tests on decided papers can draw on memory. Claude Opus 5.5 names the true first author of 27 decided ICLR 2026 papers from their titles alone, with no wrong name, and Claude Opus 4.6 accepted all 11 decided papers it said it recognized, against 5 expected. Thresholds on LLM-reviewer scores therefore have to be validated against human decisions for each model version; our tests against the ICLR 2027 decisions follow on 16 December 2026.

%% file: sec_intro.tex
\section{Introduction}
\label{sec:intro}

LLM reviewers are entering peer review. AAAI-26 gave every main-track submission an AI-generated review alongside the human ones \citep{biswas2026aaai}, ICLR 2025 offered optional LLM feedback on more than 20{,}000 randomly selected reviews \citep{thakkar2025feedback}, and a study of more than 125{,}000 paper-review pairs from ICLR, NeurIPS and ICML compares LLM-assisted reviews with the others \citep{sharma2026favor}. At ICML 2026, 22.5\% of surveyed reviewers under a no-LLM policy reported using one anyway \citep{kim2026icml}. One recent proposal maps reviewers' written rationales to LLM anchor scores and flags large gaps between human and anchor scores for escalation \citep{chen2026calibrated}; another fits model-specific acceptance thresholds on LLM scores \citep{yang2026publications}. Each of these uses reads an LLM score as a measurement of the paper.

Two properties of deployed frontier models bear on that reading. Providers replace models within months (Anthropic released five Opus versions between February and September 2026), so a score or threshold fitted to one version has to hold for the next \citep{zhang2026versionshift}. And models recognize decided papers \citep{konan2026paperscorer}, and recent Claude models name their first authors from the title (\cref{sec:boundaries}), so evaluations on decided papers can draw on memory. The score comparisons in \cref{sec:related} use papers whose decisions are public, and one error-detection study chooses papers that postdate the models' training \citep{alharbi2026scrutinise}. Version comparisons of LLM judges come from other judging tasks \citep{zhang2026versionshift,sunkavalli2026raters}.

We audit twenty models from three providers, released between February and November 2026, on ICLR 2027 submissions between 8 and 15 November 2026, after the initial human reviews and before the decisions, with web access disabled. All models review the same stratified sample of 24 submissions with the official review form, and paired comparisons extend to 57 and 60 further submissions and to 183 and 60 randomly drawn ones. We then ask where the verdict changes when the version changes, with controlled variations of the prompt and the client, a cross-scoring design that separates writing a critique from scoring it, title-only memory probes, and an edit that removes authors' statements of limitations. Four prospective plans against the ICLR 2027 decisions, covering up to 1{,}000 randomly drawn submissions, are registered and timestamped before the decisions, and their decision-free tests are already scored (\cref{sec:prereg}).

Our findings:
\begin{itemize}
\setlength\itemsep{1pt}
\item \textbf{Drift.} Between an earlier and a later version of Claude Opus and of Gemini, the accept share falls by 52 and 53 points on 183 and 60 random submissions in a registered test, and by 43 to 79 percentage points in six further paired comparisons of the same two version pairs, with no paper moving up; re-runs move 0 to 5 papers (\cref{sec:drift,sec:locate,sec:prereg}).
\item \textbf{Location.} The drift persists under three prompt and client changes. Cross-scoring splits each version gap additively into a critique, a scorer and a verdict term; the critique term is $-51$ points in the Gemini line and $-35$ in the Opus line, next to an Opus verdict term of $-30$. For Opus it is $-35$ points with GPT-6.1 Sol as a third-party scorer, $-24$ after a neutral rewrite of the critiques, and $-36$ when the scorer also reads the paper ($-47$ without it on the same papers). The same critiques receive an accept share 17 points higher from Opus 5.5 than from Opus 4.6 when the scorer sees only the critique and 14 points lower when it also sees the paper (\cref{sec:locate}).
\item \textbf{Ranking.} On 57 second-batch papers, the percentiles of Opus 4.6 and 5.5 correlate at 0.73 while their accept shares differ by 44 points. Reweighted to the population of three-review submissions, the percentiles of three models correlate with the mean of two human reviewers more strongly than the third reviewer does on 99 to 119 submissions, by 0.17 to 0.21 (point estimates; 95\% intervals between $-0.04$ and $+0.42$), and by 0.22, 0.18 and 0.20 with percentiles coarsened to the four human rating levels (Opus 5.5 interval $[+0.02,+0.42]$; \cref{sec:ranking}).
\item \textbf{Scale.} All twenty models give a 2 or a 3 in 473 of their 475 reviews, against 60 of 72 human reviews, and two GPT models give a 1 or 2 in 34 of 40 reviews of award papers they did not recognize (\cref{sec:compression}).
\item \textbf{Memory.} From titles alone, Opus 5.5 names the true first author of decided papers in all 27 cases where it gives a name. Opus 4.6 says it recognizes papers first posted months after its release and accepts all 11 decided papers it says it recognizes, against 5 expected (\cref{sec:boundaries}).
\end{itemize}

For deployment, these results argue for fixing and reporting the model version and for validating any score threshold against current human decisions for each version. For evaluation, they argue for papers whose decisions postdate the model's training and for analysis plans registered before the decisions.

%% file: sec_setup.tex
\section{Setup}
\label{sec:setup}

\paragraph{Papers.}
The main sample consists of 24 ICLR 2027 submissions with exactly three initial reviews and an arXiv version, drawn at random from three strata of the mean initial rating (low $\le 1.67$, middle $2.00$ to $2.33$, high $\ge 2.67$; 8 each) in the public OpenReview scores collected on 6 November 2026, one day after the reviews were released. A second batch of 119 submissions (120 drawn, 30 from each of four strata of the same snapshot, 119 with a retrievable PDF) is used for replication and for the prospective tests (\cref{sec:prereg}), and 60 of them (15 per stratum) are also reviewed by the endpoint versions for paired comparisons. We add 24 papers that received best, outstanding or best-student paper awards at 2025 to 2026 venues (ICLR, ICML, NeurIPS, CVPR, ICCV, ECCV, ACL, COLM, AAAI, IJCAI, KDD, SIGIR, WWW, CHI), reviewed for a top venue in their own field, and a calibration set of 40 decided ICLR 2026 main-track papers (8 each of orals, posters, rejections with mean rating $\ge 5$, rejections with mean rating $\le 3$, and withdrawals). Across the three samples, 1{,}144 of the 3{,}997 submissions we tried (29\%) had an arXiv version matching the title. In the random search of \cref{sec:prereg}, matched submissions have higher initial ratings (mean 2.19 against 2.01) and a different mix of areas (total variation 0.14, at most 0.11 over random splits); the stratum weights of \cref{sec:ranking} correct for the difference in rating. We report only aggregate results for papers under review.

\paragraph{Models.}
Twenty models: GPT-5.6 Luna, Terra and Sol; GPT-6 Luna, Sol and Astra; GPT-6.1 Terra and Sol; Gemini 3.8 Flash; Gemini 4 Argon, Carbon and Hydrogen; Claude Opus 4.6, 4.7, 4.8, 5 and 5.5; Sonnet 5.5; Haiku 5.5; Fable 5.2. They run through their providers' command-line clients (Codex CLI, Antigravity CLI, Claude Code) with web search and file tools disabled, at one reasoning-effort setting per family (OpenAI medium, Google high, Anthropic xhigh, Fable 5.2 high); Claude Opus 5.5 at medium effort gives the same rating as at xhigh on 21 of 24 papers (\cref{app:robust}). All reviews analyzed here were generated between 8 and 15 November 2026.

\paragraph{Review protocol.}
Each model receives the paper's main text (authors and acknowledgements removed, cut at the references, truncated to 60{,}000 bytes) after the ICLR 2027 review form: three sub-scores (1 to 3), strengths, weaknesses, questions, a rating from 1 (clear rejection) to 4 (clear acceptance) and a confidence, followed by a JSON line. The instruction includes one calibration sentence, ``most submissions are not accepted, so do not inflate scores, and do not deflate them either''. In the same conversation (for GPT-6.1 Sol, a fresh session with the paper and its own review) the model then reports the paper's percentile among all submissions and its probability of acceptance. \Cref{app:prompts} gives every prompt verbatim. With the full text and no truncation, 67 of 68 ratings by three models stay the same (\cref{app:robust}).

\paragraph{Recognition probes.}
In fresh sessions, each model answers a title-only probe (has it seen the paper, with guesses of first author, venue, acceptance, award and popularity) and a recall probe (authors and specific findings). A separate memory test applies the title-only probe to 60 decided ICLR 2026 titles (15 each of orals, posters, rejections and withdrawals), 30 ICLR 2027 submissions and 20 invented titles. A paper is excluded for a model if the model recalls it, or recognizes it together with its acceptance, award or popularity; on decided papers any recognition excludes it. Reviews that self-report recognition are also excluded.

\paragraph{Measures.}
The accept share is the fraction of a model's reviews rated 3 or 4. Rank agreement is the Spearman correlation between a model's percentiles and human ratings. For paired version comparisons we report the difference in accept share over papers both versions reviewed, with a 2{,}000-draw paper bootstrap, and the number of papers that cross the threshold in each direction. Weaknesses and strengths are split into items and labeled with 14 weakness and 11 strength categories by GPT-6.1 Sol; a second labeler, GPT-6 Astra, agrees at Cohen's $\kappa=0.85$ on 400 weakness items (\cref{app:robust}).

%% file: sec_results.tex
\section{Scale, ranking and drift}
\label{sec:results}

\subsection{Ratings collapse onto the two middle points}
\label{sec:compression}
The 72 human reviews of the main sample spread over the whole scale (rating 1/2/3/4: 14\%/56\%/28\%/3\%), against 18.9\%/57.5\%/22.1\%/1.5\% for all 125{,}031 initial ICLR 2027 reviews. The twenty models used a 1 or a 4 in 2 of their 475 reviews (six models in \cref{fig:dist}, \cref{app:more}). On the second batch, the four primary models (GPT-6.1 Sol, GPT-6 Astra, Gemini 4 Argon, Claude Opus 5.5) used an extreme rating in 7 of their 379 reviews (119, 119, 39 and 102; 1.8\%), against 19.9\% of 357 human reviews. GPT-6.1 Sol and GPT-6 Astra rated 23 of the 24 main-sample papers a 2, and 87\% and 90\% of the second batch. Which middle point a model settles on depends on its version. Claude Opus 4.6 rated 18 of 22 papers a 3, and Opus 5.5 rated 21 of 24 a 2.

The same two ratings cover papers the community judged exceptional. Among the award papers a model did not recognize, GPT-6.1 Sol rated 16 of 20 at 1 or 2, GPT-6 Astra 18 of 20, Gemini 4 Argon 6 of 16, Claude Opus 5.5 7 of 12 and Fable 5.2 5 of 10, and 6 of the 12 award papers that all four primary models reviewed received a 1 or 2 from each of them. Both GPT models also rated 6 of the 7 unrecognized orals of the ICLR 2026 calibration set at 1 to 2, and their median percentiles for award papers are within one point of those for the main sample (56.5 against 56.5; 59 against 58). Claude Opus 5.5 places award papers at a median of 51 against 45, and Gemini 4 Argon at 77 against 58. For the GPT models, a 2 is the usual rating for strong and weak papers alike. Both give a soundness sub-score of 1 in 22\% to 24\% of their reviews and never the top score of 3, against soundness 1 in 7\% to 8\% of reviews for Claude Opus 5.5 and Gemini 4 Argon, and every review with soundness 1 carries a rating of 1 or 2 (\cref{app:robust}).

\subsection{Rankings compared with a third reviewer}
\label{sec:ranking}
Across all 36{,}464 active ICLR 2027 submissions with three public reviews, one reviewer's rating correlates with the mean of the other two at a Spearman of 0.20. For a matched comparison we draw one human per paper as the single reviewer, take the mean of the other two as the reference, and correlate the single human and each model's percentile with that reference on the same papers. Both samples are stratified by the mean human rating, so we weight each paper by the population share of its stratum and resample papers within strata (2{,}000 draws). On the second batch, Claude Opus 5.5 reaches 0.44 against 0.23 for the single human on the same 99 papers, and GPT-6.1 Sol and GPT-6 Astra reach 0.31 and 0.32 against 0.14 on 119 papers. The paired differences, medians over draws, are $+0.21$ ($[-0.01,+0.42]$), $+0.17$ ($[-0.04,+0.37]$) and $+0.17$ ($[-0.04,+0.36]$). With model percentiles coarsened to the four-level distribution of human ratings they are $+0.22$ ($[+0.02,+0.42]$), $+0.18$ and $+0.20$ (\cref{app:more}). On the main sample the single human reaches 0.25 and the twenty models 0.13 to 0.53 (median 0.35), with paired differences from $-0.11$ to $+0.28$ (median $+0.10$, above zero for 16 of the 20; \cref{fig:ranking}); against the three-human mean without weights they reach 0.16 to 0.52 (\cref{tab:ladder}). \citet{jiang2025agentic} report a similar comparison on ICLR 2025, with 0.42 between their reviewer and one human and 0.41 between two humans. On the decided ICLR 2026 set, GPT-6.1 Sol and GPT-6 Astra separate accepted from rejected papers at AUC 0.49 and 0.51 with their ratings and at 0.66 and 0.72 with their percentiles, a paired gain of 0.17 ($[0.05,0.30]$) and 0.20 ($[0.09,0.32]$; computed before rounding). Gemini 4 Argon reaches 0.84 with its ratings and 0.92 with its percentiles. The registered tests of \cref{sec:prereg} repeat this comparison on the ICLR 2027 decisions.

\subsection{The accept share drifts across versions}
\label{sec:drift}
\begin{figure*}[t]
\centering
\includegraphics[width=0.88\textwidth]{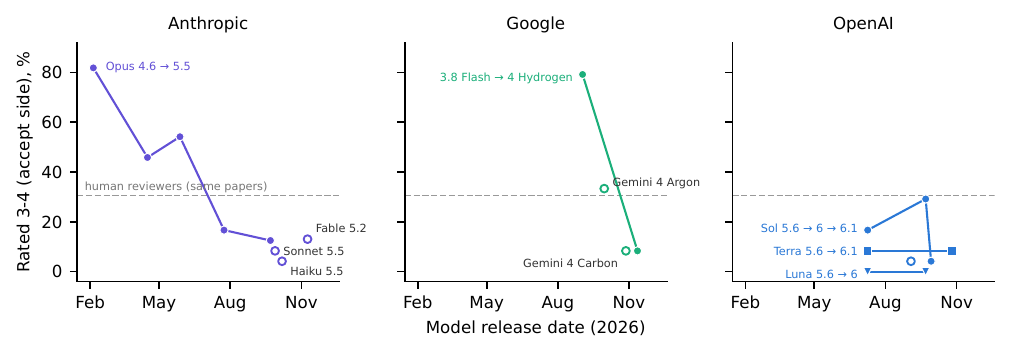}
\caption{Share of the main-sample papers rated 3 or 4 by each model (22 to 24 papers after exclusions), against release date (2026). Lines connect successive versions within a product line (for Google, Gemini 3.8 Flash and Gemini 4 Hydrogen); open circles are single models, and the open circle in the OpenAI panel is GPT-6 Astra. Dashed: share of human reviews of the same papers rated 3 or 4.}
\label{fig:drift}
\end{figure*}

On identical papers and prompts, the accept share changes by up to 71 points between versions of a product line (\cref{fig:drift}, \cref{tab:ladder}). Claude Opus 4.6 (February 2026) placed 82\% of the main sample on the accept side, Opus 4.7 46\%, Opus 4.8 54\%, Opus 5 17\% and Opus 5.5 (September 2026) 12\%. Gemini 3.8 Flash placed 79\%, and the three Gemini 4 models 33\% (Argon), 8\% (Carbon) and 8\% (Hydrogen). Human reviewers placed 31\% of their reviews of these papers on the accept side.

Paired by paper, Opus 4.6 to Opus 5.5 is $-68$ points (95\% CI $[-86,-50]$; 15 of 22 papers down, none up), and Gemini 3.8 Flash to Gemini 4 Hydrogen is $-71$ points ($[-88,-54]$; 17 down, none up). Both replicate on further second-batch papers, with $-44$ points on 57 papers for Opus ($[-56,-32]$; 25 down, none up) and $-55$ points on 60 papers for Gemini ($[-67,-43]$; 33 down, none up), with human reviews of those papers at 30\% on the accept side. Together with the two prompt changes in \cref{sec:locate} ($-79$ and $-43$), these form six paired comparisons of the two version pairs, four of them on the main-sample papers. Re-running each endpoint model on the main sample moved 1 (Opus 4.6), 1 (Opus 5.5), 5 (Flash; 2 up, 3 down) and 3 (Hydrogen; 1 up, 2 down) papers across the threshold. Keeping the reviews that self-report recognition moves each paired drift by at most one point, and giving every missing or excluded review the least favorable rating leaves the smallest drift at $-38$ points (\cref{app:robust}).

OpenAI models stay between 0\% and 29\% in every generation, and the Opus line rises once, from 46\% under Opus 4.7 to 54\% under Opus 4.8. Of the eleven comparisons between consecutive generations on the main sample (four steps of the Opus line, two of GPT Sol, one each of GPT Terra and Luna, and Gemini 3.8 Flash against each Gemini 4 model), six lower the accept share by 25 to 71 points with no paper moving up, and five stay within 12 points (\cref{app:more}).

Percentiles from different Opus versions correlate at a median Spearman of 0.77 on the main sample (range 0.47 to 0.89), and those of Opus 4.6 and 5.5 at 0.73 on the 57 second-batch papers, against 0.87 and 0.94 between two runs of the same Opus version; Flash and Hydrogen correlate at 0.44 on the main sample and 0.77 on the 60-paper comparison. Within the OpenAI lines the agreement is 0.57 to 0.80 for Sol, 0.40 for Terra and 0.10 for Luna (\cref{app:more}).

%% file: sec_locate.tex
\section{Where the drift comes from}
\label{sec:locate}

We locate the drift on the Opus line, where every intermediate version is available, and repeat the cross-scoring design on the Gemini and GPT lines.

\subsection{The version gap under prompt and client changes}
\begin{figure*}[t]
\centering
\includegraphics[width=0.82\textwidth]{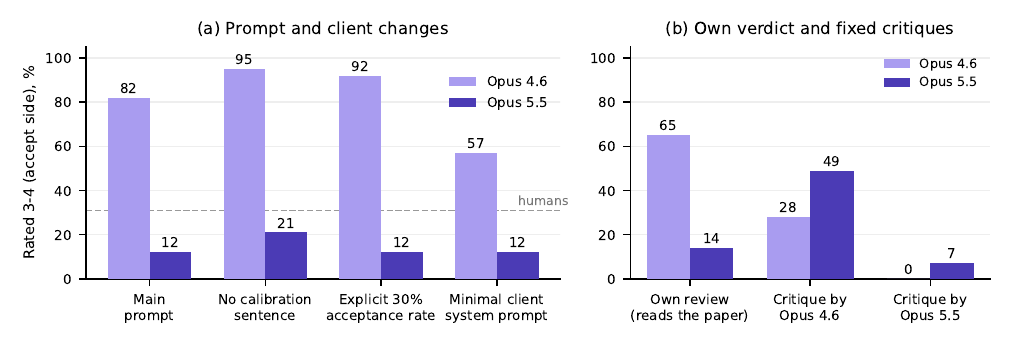}
\caption{Locating the Opus 4.6 to Opus 5.5 drift. (a) Accept share on the main sample under the main prompt and three changes to the prompt or client (Opus 4.6 on 22, 22, 24 and 23 papers after exclusions, Opus 5.5 on 24). (b) Accept share on 79 papers of each version's own review, and of score-stripped critiques written by one version and rated by either version without the paper.}
\label{fig:decomp}
\end{figure*}

Three changes test whether the instruction or the client produces the drift (\cref{fig:decomp}a). Removing the calibration sentence raises the accept share of each of the four endpoint models by 8 to 21 points and leaves the version gap at 74 points for Opus (95\% against 21\%) and 63 points for Gemini (92\% against 29\%). Replacing the sentence with an explicit acceptance rate (``about 30\% of submissions are accepted; rate so that roughly the top 30\% receive a 3 or 4'') leaves Opus 5.5 at 12\% and moves Opus 4.6 to 92\%, a paired difference of $-79$ points ($[-92,-62]$; 19 papers down, none up). Replacing the Claude Code system prompt with a one-line prompt lowers Opus 4.6 to 57\%, leaves all 24 Opus 5.5 ratings unchanged, and keeps a paired difference of $-43$ points ($[-65,-26]$; 10 down, none up). Across these changes a given model's accept share moves by at most 25 points, while the gap between versions stays between 43 and 79 points.

\subsection{Critique, scorer and verdict}
\label{sec:cross}
A review involves reading the paper and writing a critique, and then turning the critique into a rating. To separate the two stages we strip every score, sub-score and sentence mentioning acceptance, rejection or a recommendation from the reviews that two versions wrote of the same paper, and ask each version, in a fresh session, to rate each stripped critique on the same scale (\cref{app:prompts}). In the main design the scorer sees only the critique; in a second design it also sees the paper. We ran the critique-only design on one version pair per product line, for the Opus endpoints on the 22 main-sample papers (two repeats, which agree in 85 of 88 cells) and again on 57 second-batch papers (one repeat), for the Gemini and GPT pairs on the 24 main-sample papers (two repeats; 91 of 96 and 94 of 96 cells agree), and the paper-visible design on the Opus and GPT pairs, with decision thresholds for these replications written down before their calls (\cref{app:robust}). Stripping removes a median of one sentence from Opus 4.6 critiques and none from Opus 5.5 critiques.

The gap between two versions' reviews splits additively into three terms (\cref{tab:cross}; \cref{fig:decomp}b shows the pooled Opus cells). The critique term is the change in accept share when the later version's critique replaces the earlier one, averaged over the two scorers. The scorer term is the change when the later version scores the same critique, averaged over the two critiques. The verdict term is the difference between the two versions in how far each one's own review exceeds its rating of its own critique. Without the paper, the critique term is $-35$ points on the 79 Opus papers ($[-44,-27]$), $-51$ on the Gemini pair ($[-68,-34]$) and $-4$ on the GPT pair, where both versions place 2\% to 8\% of all stripped critiques on the accept side. Read without the paper, a critique receives at least as high an accept share from the later version (scorer terms $+14$, $+1$ and $0$ points), and the verdict terms are $-30$, $-21$ and $-21$ points. The terms add up by construction; the interaction of critique and scorer is within 6 points for Gemini and GPT and $-20$ and $-12$ points for Opus (\cref{app:robust}). Rating the same stripped Opus critiques twice, GPT-6.1 Sol places 35\% and 38\% of the Opus 4.6 critiques and 1\% of the Opus 5.5 critiques on the accept side, a critique term of $-35$ points on the 79 papers ($[-47,-25]$), the same as with the two Opus versions as scorers, so the effect persists when no version scores its own critiques \citep{panickssery2024recognize}. After every critique is rewritten into a neutral template that keeps each technical point and drops evaluative wording, $-24$ points of this term remain ($[-34,-16]$). The rewritten Opus 5.5 critiques list a median of 27 weaknesses per paper against 12 for Opus 4.6. GPT-6.1 Sol checked these weaknesses against the paper text on 30 papers and found 83\% of the Opus 5.5 weaknesses and 87\% of the Opus 4.6 weaknesses supported by the text, and 5\% and 8\% contradicted by it (\cref{app:robust}).

\begin{table}[t]
\centering
\small
\caption{Cross-scoring decomposition in percentage points; Gap $=$ Verdict $+$ Critique $+$ Scorer up to rounding. Reviews: accept share of each version's own review (earlier/later). Own: accept share when each version rates its own stripped critique. Pairs: Opus 4.6 and 5.5 (57: second batch; 79: both sets), Gemini 3.8 Flash and 4 Hydrogen, GPT-6 Sol and GPT-6.1 Sol.}
\label{tab:cross}
\setlength{\tabcolsep}{2.4pt}
\begin{tabular}{lccrrrr}
\toprule
Pair, $n$ & Reviews & Own & Gap & Verdict & Critique & Scorer \\
\midrule
\multicolumn{7}{l}{\emph{Critique only}} \\
Opus, 22 & 82/14 & 36/7 & $-68$ & $-39$ & $-47$ & $+17$ \\
Opus, 57 & 58/14 & 25/7 & $-44$ & $-26$ & $-31$ & $+13$ \\
Opus, 79 & 65/14 & 28/7 & $-51$ & $-30$ & $-35$ & $+14$ \\
Gemini, 24 & 79/8 & 56/6 & $-71$ & $-21$ & $-51$ & $+1$ \\
GPT, 24 & 29/4 & 8/4 & $-25$ & $-21$ & $-4$ & $0$ \\
\midrule
\multicolumn{7}{l}{\emph{Paper and critique}} \\
Opus, 22 & 82/14 & 64/14 & $-68$ & $-18$ & $-36$ & $-14$ \\
GPT, 24 & 29/4 & 38/4 & $-25$ & $+9$ & $-21$ & $-12$ \\
\bottomrule
\end{tabular}
\end{table}

With the paper in view, the verdict term shrinks from $-39$ to $-18$ points for Opus and turns from $-21$ to $+9$ for GPT. Rating its own critique together with the paper, Opus 4.6 places 64\% of the 22 main-sample papers on the accept side, against 36\% from the critique alone and 82\% in its own review, and GPT-6 Sol places 38\% against 8\% and 29\%. With the paper in view, the critique term is $-36$ points for Opus ($[-52,-20]$; $-47$ without the paper on the same papers) and $-21$ for GPT ($[-33,-8]$; $-4$ without it), and the later version becomes the harsher scorer by 14 ($[-23,-5]$) and 12 points ($[-21,-4]$). Here the interaction is $+27$ points for Opus ($[+9,+45]$; critique term $-50$ under Opus 4.6 as scorer and $-23$ under Opus 5.5) and $+17$ for GPT ($[0,+38]$). Without the paper, the critique term exceeds the verdict term by 30 points on the Gemini pair ($[-57,-1]$), and on the Opus pair the two differ by 5 points ($[-22,+11]$); with the paper, the Opus critique term is twice the verdict term ($-36$ against $-18$). \citet{zhang2026versionshift} code paired rationales of open-weight judge versions with a third-party model and find that on two of three chains most score changes of at least one point co-occur with changed cited defects; swapping the critiques themselves between versions measures the effect of the critique text directly on peer reviews. Anchor scores computed from written rationales \citep{chen2026calibrated} rest on the same step of rating a written critique. On the 22 main-sample papers, the same critiques receive an accept share 17 points higher from Opus 5.5 than from Opus 4.6 when the scorer sees only the rationale ($[+9,+26]$; $+14$ on all 79 papers) and 14 points lower when it also sees the paper ($[-23,-5]$), so this step depends on the version that performs it.

The content of the critiques changes with the version and the family (3{,}002 weakness items from the main sample; \cref{fig:heatmap} in \cref{app:more}). From Opus 4.7 on, Claude models raise novelty in 46\% to 83\% of reviews, against 18\% for Opus 4.6 and 0\% to 21\% for GPT and Gemini models. Concerns about significance fall from 41\% to 21\% of reviews between Opus 4.6 and 5.5 and from 42\% to 8\% between Gemini 3.8 Flash and Gemini 4 Hydrogen. Gemini 4 models report mathematical or technical errors in 33\% to 62\% of reviews, GPT models in 8\% to 29\% and Opus 4.6 to 4.8 in none, and GPT models say that claims exceed the evidence in 50\% to 79\% of reviews against 17\% to 33\% for Gemini models. In the 678 reviews of the four primary models on all paper sets, evaluation-method problems are, within each model, 13 to 20 points more frequent in reviews rated 1 or 2 than in reviews rated 3 or 4, and limited scope is 6 to 27 points more frequent in reviews rated 3 or 4.

\subsection{Thresholds differ between versions}
For each of the twelve models in \cref{app:more}, some percentile cut separates its ratings of 3 from its ratings of 2 on every paper, which defines an implied threshold for each version. Opus 4.6 rates a paper 3 above its own 42nd percentile, later Opus versions above their 52nd to 58th, and for every one of the twelve models the acceptance probability asked in the same follow-up separates the two ratings at a cut between 30 and 39. Gemini 3.8 Flash and Gemini 4 Hydrogen use similar thresholds (60th and 62nd), and the drift on that line comes from where they place papers (median percentile 75 against 56.5). The two Opus endpoints also place papers apart when the rating is not asked. In a fresh session that requests only a percentile and an acceptance probability, Opus 4.6 and 5.5 give values that correlate with their original percentiles at 0.78 and 0.90, and their medians differ by 24 points (65 against 41; \cref{app:robust}). A low threshold also amplifies differences between paper sets. All five models reviewed on both samples place our ICLR 2027 main sample 9 to 20 percentile points higher than the decided ICLR 2026 sample, and for Opus 4.6 this shared shift turns into an accept share of 82\% on ICLR 2027 submissions against 28\% on the ICLR 2026 papers it did not claim to recognize (\cref{app:more}). The Opus version gap varies with the paper set accordingly, from 68 points on the main sample and 44 on the second batch to 14 on the 21 decided ICLR 2026 papers that neither version said in its review it recognized (\cref{app:more}). A version change also moves the training cutoff; within ICLR 2027 the gap stays between $-44$ and $-53$ points across arXiv posting months (\cref{app:more}).

\subsection{Memory and stated limitations}
\label{sec:boundaries}
\paragraph{Memory.}
Given only the title, Claude Opus 5.5 recognized 28 of 60 decided ICLR 2026 papers (orals 10 of 15, posters 8, rejections 6, withdrawals 4) and named the true first author in all 27 cases where it gave a name; Fable 5.2 recognized 24 and named all 24 first authors correctly. Neither recognized any of 20 invented titles, and Opus 4.6 recognized 1 of the 60. GPT-6.1 Sol and GPT-6 Astra recognized none of the 60 titles, Gemini 4 Argon and Carbon 1 and 3. Of the 30 ICLR 2027 submissions in the same test, Opus 5.5 recognized 2 and Fable 5.2 1. In the probes run before reviewing, Opus 5.5 flagged 24 of the 40 calibration papers, including all 8 orals, and named the true first author in all 22 cases where it gave a name, and Fable 5.2 flagged 13 of the 23 award papers it was probed on.

In its reviews of ICLR 2027 submissions, Opus 4.6 stated in 19 of 284 cases, 14 of them in the 200 random submissions of plan 3, that it recognized the paper, and it placed 15 of them on the accept side; all 19 first appeared on arXiv three to nine months after its release. Opus 5.5 made no such statement in 323 reviews of ICLR 2027 submissions. On the 39 decided ICLR 2026 papers that both Opus endpoints reviewed, the probes of Opus 4.6 flagged none, yet Opus 4.6 stated in 11 reviews that it recognized the paper and accepted all 11, among them three rejected papers and one withdrawn paper. Its acceptance rates on papers of the same decision class that it did not claim to recognize predict 5.1 acceptances for these 11. Allowing for the uncertainty of these class rates, each estimated from 4 to 8 papers, all 11 would be accepted with probability $6\times10^{-4}$. Opus 5.5 stated recognition in 14 reviews and accepted 5, against 5.0 predicted in the same way. Recognition stated inside a review thus goes with acceptance in Opus 4.6 and leaves acceptance at the class rate in Opus 5.5. \citet{konan2026paperscorer} report that GPT-5.5 predicts acceptance as accurately on papers it recognizes as on others. Reviews that state recognition are excluded from all other analyses (\cref{sec:setup}).

\paragraph{Stated limitations.}
LLM reviews repeat the limitations that authors disclose more often than human reviews do \citep{ye2024risks}, and from Opus 5 on, Claude models list honest reporting of limitations as a strength in 50\% to 67\% of their main-sample reviews (Opus 4.6: 18\%; 2{,}016 strength items). We test whether stating limitations changes the score. On 21 public papers that GPT-6.1 Sol, GPT-6 Astra and Claude Opus 5.5 had reviewed without recognizing, deleting the authors' statements of limitations and negative results (median 1{,}300 characters), against deleting background text of matched length, changed GPT percentiles by $-1.1$ points on average ($[-2.5,+0.1]$) and Opus 5.5 percentiles by $-1.4$ ($[-4.0,+0.8]$), and Opus 5.5 placed the same 4 papers on the accept side in both versions (\cref{app:disclosure}). Prior work finds that confident phrasing raises LLM review scores \citep{vasu2026style}; on these 21 papers, stating limitations costs a paper at most 0.1 GPT percentile points and 0.8 Opus 5.5 points on average (upper ends of the 95\% intervals).

%% file: sec_prereg.tex
\section{Prospective tests against ICLR 2027 decisions}
\label{sec:prereg}

Rankings can be checked against human scores today; whether they predict decisions can only be checked after ICLR 2027 announces them on 16 December 2026. We fixed three analysis plans, each before reviewing the new papers it covers, a fourth plan that reuses existing reviews, and the scoring script, all before any decision was public. Each file is timestamped through OpenTimestamps in the Bitcoin blockchain, the first two plans in block 975799 (14 November 2026). Plan 1 (11 November) covers 143 papers, the 24 main-sample papers as a pilot and 119 new papers from four strata, reviewed by GPT-6.1 Sol, GPT-6 Astra, Claude Opus 5.5 (102 of the 119 new papers) and, on 39 of them, Gemini 4 Argon. Plan 2 (14 November) adds 1{,}000 submissions drawn at random from all active three-review submissions with an arXiv version, reviewed by the two GPT models. Plan 3 (15 November) reviews the first 200 of these with Claude Opus 4.6 and 5.5 and the first 60 with Gemini 3.8 Flash and Gemini 4 Hydrogen. Plan 4 (16 November, block 976009) tests whether the per-version cutoffs of \cref{sec:discussion} change balanced accuracy against the decisions on the second-batch version pairs.

Plans 1 and 2 ask three questions of each model. Q1 asks whether percentiles predict acceptance as well as a single human's initial score (AUC difference within $\pm 0.05$: comparable; at least 0.10 apart: better or worse). Q2 asks whether percentiles add information beyond the mean initial human score (logistic regression with a bootstrap interval). Q3 asks whether ratings are informative (AUC $\le 0.60$: not informative; $\ge 0.70$: informative). Plan 3 asks whether the version drift holds on random submissions (Q5a, paired difference of at most $-25$ points) and whether two versions rank the decided papers equally well (Q5b, AUC difference within $\pm 0.05$).

Two registered tests need no decisions. On the submissions we reviewed, all six models analyzed in \cref{app:more} give papers first posted on arXiv from July 2026 on higher percentiles than earlier ones after controlling for the mean human rating ($+4.5$ to $+13.6$ points), and plan 2 tests this pattern on the 1{,}000 random submissions (Q4). There the residual percentiles of the later papers are 5.2 points higher from GPT-6.1 Sol ($[3.7,6.7]$) and 5.0 points higher from GPT-6 Astra ($[3.6,6.4]$), which meets the registered criterion for replication. Plan 3 tests the version drift on random submissions (Q5a). On 183 random submissions with valid reviews from both Opus versions, Opus 5.5 places 10\% on the accept side against 62\% for Opus 4.6, a paired difference of $-52$ points ($[-58,-45]$; 95 down, none up); on 60, Gemini 4 Hydrogen places 15\% against 68\% for Gemini 3.8 Flash ($-53$, $[-65,-40]$; 32 down, none up). Both differences pass the registered criterion of at most $-25$ points.

The tests that need the decisions (Q1 to Q3, Q5b and plan 4) will be reported in a revised version after 16 December 2026.

%% file: sec_related.tex
\section{Related work}
\label{sec:related}

\paragraph{LLMs as reviewers.}
The overlap between the points raised by GPT-4 and by human reviewers is comparable to the overlap between two human reviewers \citep{liang2023feedback}, and in an exploratory study GPT-4 found inserted errors in 7 of 13 short papers but erred on 6 of 10 pairs of abstracts that differed clearly in quality \citep{liu2023reviewergpt}. Automated review appears in research agents \citep{lu2024aiscientist} and in simulations and evaluation frameworks for peer review \citep{jin2024agentreview,garg2025revieweval}. On decided ICLR papers, models from three providers separate accepted from rejected papers while aligning less with finer distinctions \citep{camelo2026closely}; raw scores are lenient and compressed, and model-specific thresholds fitted on ICLR 2025 transfer to ICLR 2024 \citep{yang2026publications}; direct ratings compress most predictions above the acceptance threshold \citep{huang2026paradigms}; area-chair guidance in the prompt reduces the mean divergence from human scores from 1.46 to 0.20 points while rank correlation moves from 0.377 to 0.387 \citep{qin2026interpreting}. Other studies measure run-to-run variability \citep{yang2026believe}, predictability of acceptance with memorization probes \citep{konan2026paperscorer}, detection of injected errors \citep{zubeldia2026auditing,alharbi2026scrutinise}, pairwise designs \citep{zhang2025pairwise}, sensitivity to confident language \citep{vasu2026style} and review behavior across 1{,}000 ICLR and NeurIPS papers \citep{zurawicki2026praib}; a registered validation of a first-pass score reaches AUROC 0.82 on 300 decided ICLR submissions \citep{georgantas2026bottleneck}. An estimated 6.5\% to 16.9\% of review text at four 2023 and 2024 ML and NLP conferences could have been substantially modified by LLMs \citep{liang2024monitoring}, and current detectors cannot enforce policies that allow LLMs only for polishing reviews \citep{saha2026policies}. Authors can steer LLM reviews with hidden prompt injections \citep{ye2024risks,keuper2025injection,li2026llmreviewer}, a vulnerability linked to how models attribute text to roles \citep{ye2026roleconfusion}, and LLM reviews repeat the limitations that authors disclose more often than human reviews do \citep{ye2024risks}; a score supplied to an LLM reviewer changes the evidence its review presents \citep{preiss2026anchors}. These studies vary the input; we hold papers and prompt fixed and vary the model version. On decided ICLR 2026 papers, a five-model panel orders rejected papers in line with the venue's post-rebuttal ratings while its absolute grade depends on the panel roster \citep{lu2026grade}, and \citet{jiang2025agentic} report that their agentic reviewer agrees with a single human as closely as two humans do on ICLR 2025. \citet{chen2026calibrated} propose LLM anchor scores to flag miscalibrated human reviews, and \citet{baumann2026stop} argue against deploying AI reviews without rigorous evaluation. We add submissions under review, version ladders in three product lines, a decomposition of where the version effect arises, and registered prospective tests.

\paragraph{Judges across versions.}
The behavior of a commercial model changes across versions \citep{chen2023chatgpt}. For LLM judges, \citet{zhang2026versionshift} separate scale shifts from ranking changes across open-weight version chains and two API pairs, and by coding paired rationales find that on two of three chains most score changes of at least one point co-occur with changed cited defects; \citet{yang2026judgechanges} find that judge upgrades are not interchangeable; LLM judges carry position, verbosity and other biases under controlled perturbations \citep{ye2025justice} and favor their own generations \citep{panickssery2024recognize}; silent updates of API models can be detected from log probabilities \citep{chauvin2026logprob}; replacing a judge with one from another provider changes system rankings \citep{han2026swapping}; a fixed human-labeled anchor set can attribute a drift alarm to the judge \citep{li2026whodrifted}; \citet{sunkavalli2026raters} report severity shifts in all five version contrasts of a registered essay-scoring audit; and newer relevance judges need not preserve earlier correct judgments \citep{yu2026backbone}. As in their chains, scales move between versions; of the eleven consecutive-generation steps in our three lines, six lower the accept share by 25 to 71 points and none raises it by more than 12. Implied scores inferred from score-stripped human reviews differ from the assigned scores, and the offset predicts later score changes \citep{louis2026rebuttals}. An LLM reviewer can state serious weaknesses while its score stays above the decision threshold \citep{mao2026dsar}. Our verdict term is the analogous offset for model reviews, measured across versions. For peer review we split the movement by intervention into a critique, a scorer and a verdict term, and find that the percentiles of Opus 4.6 and 5.5 correlate at 0.73 on 57 papers on which their accept shares differ by 44 points.

\paragraph{Contamination.}
Contamination can be detected from model behavior \citep{golchin2023timetravel} and avoided by freezing model predictions before outcomes are known \citep{karger2025forecastbench}, and LLMs can infer the authors of anonymized papers \citep{gloire2026doubleblind,zhang2026deanon}. We show that recent Claude models recall decided papers and their first authors from titles alone, and that one version accepts every decided paper it says it recognizes. Human review is itself noisy \citep{beygelzimer2023neurips,huang2026lottery}, so we compare models with a single human reviewer.

%% file: sec_discussion.tex
\section{Discussion}
\label{sec:discussion}

\paragraph{Deployment.}
A rule that flags papers an AI rates below 3 would have flagged 4 of 22 main-sample papers under Opus 4.6 and 21 of 24 under Opus 5.5, and with an explicit acceptance rate in the prompt the two versions still differ by 79 points. A venue that uses LLM scores therefore needs to fix the model version and report it with the client and reasoning effort. Any threshold then has to be validated against current human decisions for each version, and re-scoring a fixed human-labeled anchor set with each new version can attribute such a change to the judge \citep{li2026whodrifted}. Percentile cutoffs set per version are one way to re-anchor. On the second batch, with each version taking its top 30\% by its own percentile, the versions agree on 46.5 of the 56 Opus papers with percentiles from both versions and 50.0 of 60 Gemini papers (12.3 of 17 and 13.0 of 18 shared selections; means over random tie-breaks), against 32 and 27 for raw ratings on the same papers and 32.3 and 34.8 expected by chance. Relative to raw ratings, these cutoffs change balanced accuracy against the initial human mean by $-0.07$ to $+0.03$, every interval including zero (\cref{app:robust}), and the same comparison against the final decisions was registered before their release (plan 4, \cref{sec:prereg}). Anchor scores computed from written rationales \citep{chen2026calibrated} need the same re-anchoring, since rating a written critique, the step they rest on, moves by 14 to 17 points between two versions of one model, in a direction that depends on whether the scorer sees the paper.

\paragraph{Finding faults and weighing them.}
GPT-6.1 Sol finds 83\% to 87\% of the Opus weaknesses supported by the paper text (\cref{sec:cross}), and on five award papers whose reviews we compared with the full text, the specific problems the models raised pointed to identifiable passages. On one ICML 2026 outstanding paper, all four primary models raised the same discrepancy between the abstract and a later section and rated the paper 4, 4, 3 and 2. Models and versions differ in which problems they raise and in how they weigh them (\cref{sec:cross}). This favors using LLMs to surface candidate issues and to rank papers, and leaving the weighing of the issues to people.

\paragraph{Evaluation.}
Evaluations of LLM reviewers should report the ranking signal and the scale separately. Designs that place a paper among reference papers of known standing, or compare papers in pairs \citep{zhang2025pairwise}, build on the part of the signal that survives a version change. A version change moves the Opus accept share by 43 to 79 points across the prompts we tested, while the percentiles of Opus 4.6 and 5.5 correlate at 0.73 on 57 second-batch papers. Since part of the version effect sits in the critique, comparisons of critique content across studies depend on the version as well. Recent Claude models recall decided papers, including public rejections and withdrawals, and name their first authors, and an older version states recognition of papers first posted after its release, so a retrospective benchmark needs recognition probes and a check of the reviews themselves. Papers whose decisions postdate the model's training, as in our registered tests, rule out that the model knows the decision.

\paragraph{Limitations.}
The version ladder covers up to 24 papers per model, with paired replications on 57 and 60 papers and on 183 and 60 random submissions for the Opus and Gemini endpoints. The cross-scoring decomposition covers one version pair in each of the three lines. Samples are drawn from three-review submissions with an arXiv version, which differ from the rest in rating and in area (\cref{sec:setup}), and some human reviews in the reference may be AI-assisted \citep{latona2024lottery,kim2026icml}. Models were accessed through their providers' clients at one reasoning-effort setting per family, and all reviews use the ICLR 2027 review form. The decision-based tests will be scored after 16 December 2026.

%% file: sec_statements.tex
\section*{Impact Statement}
This work concerns the use of LLMs in peer review. All model reviews were generated privately from the public arXiv versions of the papers, no non-public review material was sent to any model, and no model output was posted to OpenReview or shared with authors, reviewers or area chairs. For papers under review we report only aggregate statistics and no per-paper scores, quotes or identifiers. Award papers and decided ICLR 2026 papers are public, and we report them in aggregate, with one award paper described without its title. Human scores are public OpenReview data. The findings argue against reading LLM scores as stable measurements of paper quality, and for version control and threshold validation in any deployment. The memory results show that recent models can identify the authors of decided papers from their titles, which matters for double-blind review of resubmitted papers. Matching submissions to arXiv versions revealed their authors to us; we did not contact authors and did not use their identities in any analysis. The prompts, analysis scripts, registration files with their timestamps and per-model aggregate outputs are available from the author.

%% file: sec_appendix.tex
\clearpage
\appendix
\onecolumn
\section{Prompts and clients}
\label{app:prompts}

\paragraph{Review prompt.} The submission text follows the last line. For award papers, the first sentence reads ``You are an experienced reviewer for a top-tier conference in this paper's own field (for example ICLR, NeurIPS, CVPR, ACL, SIGIR or CHI, whichever fits the paper).'', the form is labeled ``adapted from ICLR 2027'', and the calibration sentence refers to ``a typical reviewer at that venue''.
\begin{lstlisting}
You are an experienced reviewer for ICLR 2027. Below is the main text of a submission (author names, affiliations, references and appendix removed). Write a review using the ICLR 2027 review form below. Be calibrated like a typical ICLR 2027 reviewer: most submissions are not accepted, so do not inflate scores, and do not deflate them either.

Rely only on the text provided. Do not search the web, do not use any tools, and do not read any files.

Review form:
Paper Summary:
Presentation (1-3): 3 = Clear and easy to follow; 2 = Mostly clear, some parts hard to follow; 1 = Difficult to follow despite careful reading
Informativeness (1-3): 3 = I learned a lot / substantial new insight; 2 = I learned something interesting or useful; 1 = I learned little or nothing new from this paper
Soundness (1-3): 3 = No significant concerns; 2 = I have minor or fixable concerns about the paper's soundness; 1 = I have significant concerns about the paper's soundness that do not appear easy to address
Critical Strengths:
Critical Weaknesses:
Questions For Authors:
Rating (1-4): 1 = Clear rejection; 2 = Weak rejection; 3 = Weak acceptance; 4 = Clear acceptance
Confidence (1-4): 1 = Educated guess; 2 = Willing to defend, but parts may be outside my expertise; 3 = I am familiar with this topic and its relevant literature; 4 = I have substantial direct research experience on this topic

After the review, on the very last line output exactly one JSON object and nothing after it:
{"rating": <1-4>, "confidence": <1-4>, "presentation": <1-3>, "informativeness": <1-3>, "soundness": <1-3>, "recognized": <true or false>}
where "recognized" says whether you recognize this specific paper from before reading it here.

=== SUBMISSION TEXT ===
\end{lstlisting}

\paragraph{Follow-up question} (same conversation; for GPT-6.1 Sol on the main sample and the 119-paper batch, a fresh session containing the review prompt with the paper and the model's own review). For award papers, ``accepted at ICLR 2027'' reads ``accepted at a top-tier conference in its field'' and ``all ICLR 2027 submissions'' reads ``all submissions to that conference''.
\begin{lstlisting}
Follow-up question about the submission you just reviewed. Answer from your review above only; do not use any tools or search. Give (1) your estimated probability, 0-100, that this paper is accepted at ICLR 2027, and (2) your estimated percentile of this paper among all ICLR 2027 submissions (0 = weakest, 100 = strongest). Reply with exactly one JSON object and nothing else: {"accept_prob": <integer>, "percentile": <integer>}
\end{lstlisting}

\paragraph{Title probe} (fresh session).
\begin{lstlisting}
Answer from memory only. Do not search the web, do not use any tools, do not read any files.
Have you seen a research paper with this exact title before?
Title: "{title}"
Reply with exactly one JSON object and nothing else:
{"recognized": true or false, "first_author_guess": string or null, "venue_guess": string or null, "year_guess": integer or null, "accepted_somewhere": "yes" or "no" or "unknown", "award": string or null, "popularity": 0-3 (0 = never heard of it, 1 = vaguely, 2 = known in its subfield, 3 = widely known), "confidence": number between 0 and 1}
\end{lstlisting}

\filbreak
\paragraph{Recall probe} (fresh session).
\begin{lstlisting}
Do not search the web, do not use any tools, do not read any files. Answer only from your own memory.
Please try hard to recall the research paper titled: "{title}"
Think carefully about whether you have actually read or seen this paper (for example on arXiv, OpenReview, a conference program, social media, or a blog), as opposed to guessing from the title.
If you genuinely remember it, state concretely: the authors, the specific main technical result (method, theorem, numbers, or findings that cannot be guessed from the title alone), where and when it was published, and whether it won any award. If you do not remember it, say so; do not guess.
Reply with exactly one JSON object and nothing else:
{"remembers": true or false, "authors": string or null, "main_result": string or null, "venue": string or null, "year": integer or null, "award": string or null, "how_known": string or null, "confidence": number between 0 and 1}
\end{lstlisting}

\paragraph{Clients and settings.} OpenAI models: Codex CLI 0.162.1 and 0.167.0 (\texttt{codex exec}, read-only sandbox, web search off), reasoning effort medium. Google models: Antigravity CLI in plan mode (no tools), effort suffix \texttt{-high}. Anthropic models: Claude Code (\texttt{claude -p}) with web, file and shell tools disabled, effort xhigh (Fable 5.2: high). On 14 November 2026 the Claude Code and Antigravity clients were at versions 2.1.323 and 1.5.9. Codex CLI and Claude Code log the model behind each answer; for Antigravity CLI we record the requested identifier. Model identifiers: \texttt{gpt-5.6-luna}, \texttt{gpt-5.6-terra}, \texttt{gpt-5.6-sol}, \texttt{gpt-6-luna}, \texttt{gpt-6-sol}, \texttt{gpt-6-astra}, \texttt{gpt-6.1-terra}, \texttt{gpt-6.1-sol}; \texttt{gemini-3.8-flash-high}, \texttt{gemini-4-argon-high}, \texttt{gemini-4-carbon-high}, \texttt{gemini-4-hydrogen-high}; \texttt{claude-opus-4-6}, \texttt{claude-opus-4-7}, \texttt{claude-opus-4-8}, \texttt{claude-opus-5}, \texttt{claude-opus-5-5}, \texttt{claude-sonnet-5-5}, \texttt{claude-haiku-5-5}, \texttt{claude-fable-5-2}.

\paragraph{Explicit acceptance rate.} In \cref{sec:locate}, the calibration sentence of the review prompt is replaced by: ``At ICLR, about 30\% of submissions are accepted. Rate so that roughly the top 30\% of submissions would receive a rating of 3 or 4.''

\paragraph{Minimal client system prompt.} In \cref{sec:locate}, the Claude Code system prompt is replaced by ``You are a helpful assistant.'' (option \texttt{--system-prompt}), for the review and the follow-up.

\paragraph{Cross-scoring prompt} (fresh session; the stripped critique follows the last line).
\begin{lstlisting}
You are an experienced reviewer for ICLR 2027. Below is another reviewer's assessment of a submission: a summary, the strengths, the weaknesses and the questions they raised. You do not see the paper. Based only on this assessment, give the overall rating it supports on the ICLR 2027 scale. Be calibrated like a typical ICLR 2027 reviewer: most submissions are not accepted, so do not inflate scores, and do not deflate them either.

Rating (1-4): 1 = Clear rejection; 2 = Weak rejection; 3 = Weak acceptance; 4 = Clear acceptance
Soundness (1-3): 3 = No significant concerns; 2 = Minor or fixable concerns; 1 = Significant concerns

Do not search the web, do not use any tools, and do not read any files. Reply with exactly one JSON object and nothing else: {"rating": <1-4>, "soundness": <1-3>}

=== ASSESSMENT ===
\end{lstlisting}

\paragraph{Critique rewriting} (GPT-6 Astra; the stripped critique follows the last line).
\begin{lstlisting}
Rewrite the peer-review assessment below into a fixed neutral format. Rules: (1) Output exactly four sections: 'Summary:' one factual sentence on what the paper does; 'Strengths:' a bullet list; 'Weaknesses:' a bullet list; 'Questions:' a bullet list. (2) Keep every distinct technical point of the assessment in its section and add no point; merge points that say the same thing. (3) Each bullet has at most 25 words. (4) Use neutral wording: remove intensifiers and evaluative adjectives (for example significant, major, minor, severe, serious, critical, excellent, strong, weak, novel, impressive, limited, fundamental) and remove any overall judgment, recommendation, score or statement about acceptance. (5) Do not use tools or read files. Output only the rewritten assessment.

=== ASSESSMENT ===
\end{lstlisting}

\paragraph{Weakness fact check} (GPT-6.1 Sol; the paper text and the numbered weaknesses of one rewritten critique follow).
\begin{lstlisting}
Below are a submission and a numbered list of weaknesses that a reviewer raised about it. For each weakness, judge from the submission text alone: S = the text supports the weakness as stated (the issue is present); C = the text addresses the weakness or contradicts it; U = it cannot be determined from the text. Do not use tools or search. Reply with exactly one JSON object and nothing else: {"labels": ["S", "C", ...]} with one label per weakness, in order.

=== SUBMISSION ===
\end{lstlisting}

\paragraph{Limitation-passage annotation} (GPT-6 Astra; the numbered paper text follows the last line).
\begin{lstlisting}
Read the paper text below. Its lines are numbered. Do not use any tools; answer from the text only.

Mark two kinds of passages by inclusive line ranges. Keep each range tight (1 to 6 lines) and cover only the lines that contain the sentences in question.
(A) Sentences in which the authors state a weakness of their own work: an explicit Limitations section or paragraph; statements that their method fails, underperforms a baseline, or degrades in some setting; negative or null results of their own experiments; caveats that restrict the scope or validity of their own claims (unverified assumptions, missing guarantees, small samples, settings not covered). Do NOT include: tables or numbers alone; ablations or comparisons in which their method performs well; statements of their contributions; descriptions of other work's limitations; ethics or broader-impact boilerplate.
(B) Neutral passages from the background or related-work parts that describe other work or general context, with no results of this paper, no claims about this paper's contribution, and no limitations of any work. The total number of characters in (B) should be within 20% of the total in (A). (A) and (B) must not overlap. List (B) ranges in order of preference.

Reply with exactly one JSON object and nothing else:
{"A": [[start, end], ...], "B": [[start, end], ...], "notes": "<one sentence on what (A) contains>"}
If there is no passage of kind (A), reply {"A": [], "B": [], "notes": "none"}.

=== PAPER (numbered lines) ===
\end{lstlisting}

\paragraph{Second-model filter} (GPT-6.1 Sol; the passages follow).
\begin{lstlisting}
Below are numbered passages from one research paper, each shown with a few lines of context. For each passage, classify the marked lines (between >>> and <<<):
W = the authors state a weakness of their own method or claims: a limitation, a failure or degradation, a negative or null result of their own method, or a restriction on the scope or validity of their own conclusions.
R = a main result or contribution of the paper, even if its content is negative (for example an impossibility result or a finding that existing methods fail, when establishing that is the paper's goal).
S = problem setup, definitions, notation or assumptions stated as part of the setup.
O = anything else (results tables, favorable comparisons, motivation about other work).
Do not use any tools. Reply with exactly one JSON object and nothing else: {"labels": ["W", "R", ...]} with one label per passage, in order.
\end{lstlisting}

\section{Additional results}
\label{app:more}

\begin{figure}[h]
\centering
\includegraphics[width=0.6\textwidth]{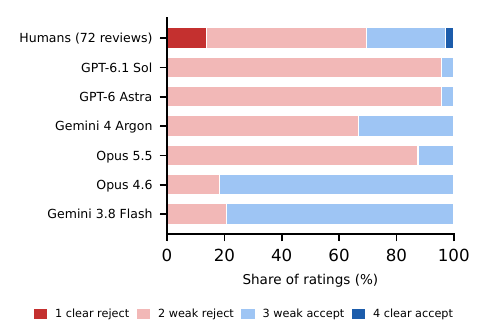}
\caption{Rating distributions on the 24 main-sample papers: human reviews and six models. Colors: 1 clear reject (dark red), 2 weak reject, 3 weak accept, 4 clear accept (dark blue).}
\label{fig:dist}
\end{figure}

\begin{figure}[h]
\centering
\includegraphics[width=0.95\textwidth]{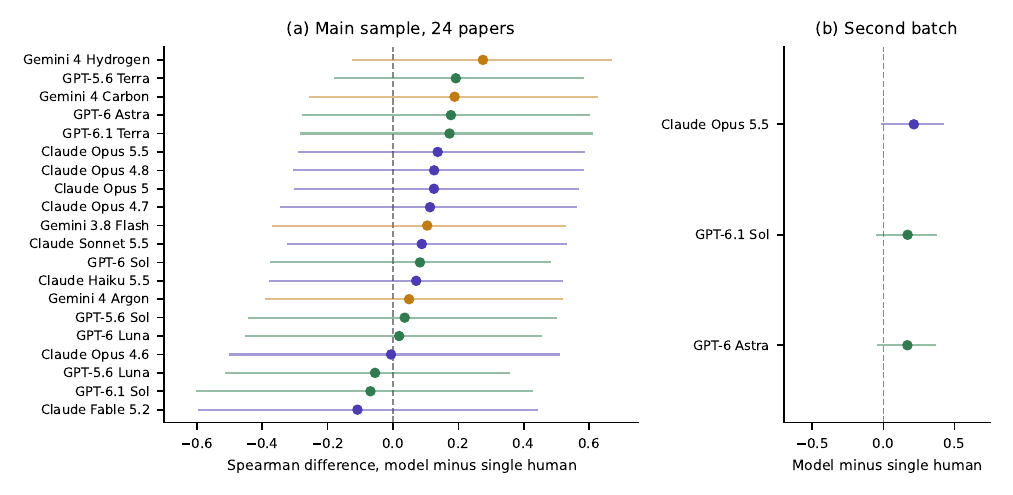}
\caption{Matched comparison with human reviewers on (a) the 24 main-sample papers for all twenty models and (b) the second batch. For each paper one human is drawn as the single reviewer and the mean of the other two is the reference; dots show the median difference between the Spearman correlation of a model's percentile with the reference and that of the single human, bars the 95\% interval (2{,}000 draws, papers resampled within strata and weighted to the population of three-review submissions). In (b), 119 papers for the GPT models and 99 for Claude Opus 5.5.}
\label{fig:ranking}
\end{figure}

\paragraph{Weighting of the matched comparison.} Both samples were drawn in equal strata of the mean of the three initial ratings, whose population shares among the 36{,}464 active three-review submissions are 0.30, 0.54 and 0.16 for the main-sample strata and 0.30, 0.30, 0.23 and 0.16 for the second-batch strata. Claude Opus 5.5 returned reviews for 102 of the 119 second-batch papers and percentiles for 99. Each paper is weighted by the population share of its stratum divided by its sample share, and the Spearman correlation is computed as the weighted Pearson correlation of weighted mid-ranks. Without weights, the single human reaches 0.33 on the main sample and 0.18 on the second batch, and the paired difference for Claude Opus 5.5 on the second batch is $+0.22$ ($[+0.01,+0.44]$).

\begin{figure}[h]
\centering
\includegraphics[width=0.85\textwidth]{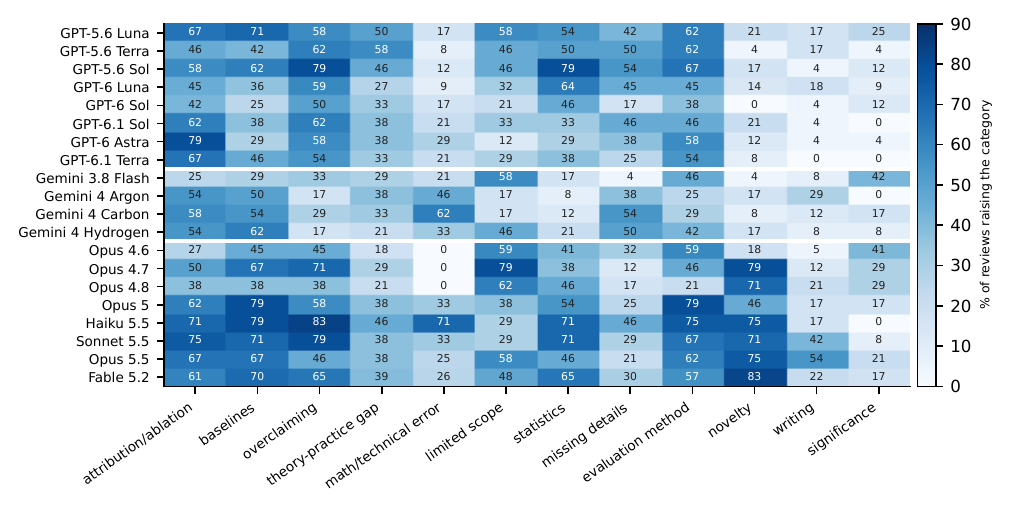}
\caption{Share of each model's reviews raising each of 12 weakness categories at least once (24 main-sample papers; labels by GPT-6.1 Sol; the categories cost and other are omitted). Claude rows list the Opus line by release, then Haiku, Sonnet and Opus 5.5, then Fable 5.2.}
\label{fig:heatmap}
\end{figure}

\begin{table}[h]
\centering
\small
\caption{Twenty models on the 24 main-sample papers. $n$: papers reviewed after exclusions. Accept: share rated 3 or 4. Pct: median self-estimated percentile. $\rho$: Spearman with the three-human mean.}
\label{tab:ladder}
\setlength{\tabcolsep}{3pt}
\begin{tabular}{llrrrr}
\toprule
Model & Release & $n$ & Accept & Pct & $\rho$ \\
\midrule
GPT-5.6 Luna & 07-09 & 24 & 0\% & 39 & 0.16 \\
GPT-5.6 Terra & 07-09 & 24 & 8\% & 63 & 0.40 \\
GPT-5.6 Sol & 07-09 & 24 & 17\% & 59 & 0.31 \\
GPT-6 Astra & 09-03 & 24 & 4\% & 58 & 0.44 \\
GPT-6 Luna & 09-22 & 22 & 0\% & 36.5 & 0.22 \\
GPT-6 Sol & 09-22 & 24 & 29\% & 65 & 0.31 \\
GPT-6.1 Sol & 09-29 & 24 & 4\% & 56.5 & 0.24 \\
GPT-6.1 Terra & 10-26 & 24 & 8\% & 60 & 0.42 \\
\midrule
Gemini 3.8 Flash & 09-02 & 24 & 79\% & 75 & 0.41 \\
Gemini 4 Argon & 09-30 & 24 & 33\% & 58 & 0.31 \\
Gemini 4 Carbon & 10-28 & 24 & 8\% & 58 & 0.43 \\
Gemini 4 Hydrogen & 11-12 & 24 & 8\% & 56.5 & 0.49 \\
\midrule
Claude Opus 4.6 & 02-05 & 22 & 82\% & 55 & 0.44 \\
Claude Opus 4.7 & 04-16 & 24 & 46\% & 55 & 0.45 \\
Claude Opus 4.8 & 05-28 & 24 & 54\% & 62 & 0.51 \\
Claude Opus 5 & 07-24 & 24 & 17\% & 51 & 0.52 \\
Claude Opus 5.5 & 09-22 & 24 & 12\% & 45 & 0.43 \\
Claude Sonnet 5.5 & 09-28 & 24 & 8\% & 45 & 0.34 \\
Claude Haiku 5.5 & 10-07 & 24 & 4\% & 30 & 0.29 \\
Claude Fable 5.2 & 11-09 & 23 & 13\% & 58 & 0.25 \\
\midrule
Human reviews & n/a & 24 & 31\% & n/a & n/a \\
\bottomrule
\end{tabular}
\end{table}

\paragraph{Cross-version rank agreement.} Spearman correlations between versions' percentiles on the main sample: Opus line median 0.77 (range 0.47 to 0.89 over ten pairs); GPT-5.6 Sol, GPT-6 Sol and GPT-6.1 Sol 0.57 to 0.80; GPT-5.6 Terra and GPT-6.1 Terra 0.40; GPT-5.6 Luna and GPT-6 Luna 0.10; Gemini 3.8 Flash and Gemini 4 Hydrogen 0.44 (24 papers) and 0.77 (60 papers); Gemini 4 Argon, Carbon and Hydrogen 0.64 to 0.88. Two runs of Opus 4.6 correlate at 0.94 and two runs of Opus 5.5 at 0.87.

\paragraph{Consecutive generations.} Paired differences in accept share on the main sample, with 95\% intervals and papers moving down and up: Opus 4.6 to 4.7 $-36$ ($[-55,-18]$; 8 down, 0 up), 4.7 to 4.8 $+8$ ($[-8,+25]$; 1, 3), 4.8 to 5 $-38$ ($[-58,-17]$; 9, 0), 5 to 5.5 $-4$ ($[-21,+8]$; 2, 1); GPT-5.6 Sol to GPT-6 Sol $+12$ ($[-4,+33]$; 1, 4), GPT-6 Sol to GPT-6.1 Sol $-25$ ($[-42,-8]$; 6, 0); GPT-5.6 Terra to GPT-6.1 Terra 0 (1, 1); GPT-5.6 Luna to GPT-6 Luna 0 (0, 0); Gemini 3.8 Flash to Gemini 4 Argon $-46$ ($[-67,-25]$; 11, 0), to Carbon $-71$ ($[-88,-50]$; 17, 0) and to Hydrogen $-71$ ($[-88,-54]$; 17, 0).

\paragraph{Coarsened percentiles.} Human ratings take four values and percentiles take integer values from 0 to 100. Mapping each model's percentiles on the second batch to four levels with the distribution of human ratings in active three-review submissions (19.1\%, 57.5\%, 21.9\%, 1.5\%) gives paired differences against the single human of $+0.22$ ($[+0.02,+0.42]$) for Claude Opus 5.5, $+0.18$ ($[-0.03,+0.38]$) for GPT-6.1 Sol and $+0.20$ ($[-0.00,+0.39]$) for GPT-6 Astra.

\paragraph{Implied thresholds.} The percentile above which a version rates a paper 3, chosen to separate its ratings best (in parentheses, papers separated): Opus 4.6 42.5 (22 of 22), Opus 4.7 55.5 (24 of 24), Opus 4.8 58.5 (24 of 24), Opus 5 58.5 (24 of 24), Opus 5.5 52.5 (24 of 24), Gemini 3.8 Flash 60.5, Gemini 4 Argon 60.5, Gemini 4 Carbon 65.5, Gemini 4 Hydrogen 62.5 (all 24 of 24), GPT-6 Sol 68.5, GPT-6.1 Sol 60.5, GPT-6 Astra 65.5 (24 of 24). The acceptance probability asked in the same follow-up separates the ratings in the same way (cut points between 30 and 39), and it concentrates near 50 for papers rated 3 and near 25 to 30 for papers rated 2.

\paragraph{ICLR 2026 against ICLR 2027 samples.} Opus 4.6 reviewed all 40 decided ICLR 2026 papers; no probe flagged any of them, and it self-reported recognition in 11 reviews (4 orals, 3 posters, 3 borderline rejections, 1 withdrawal). Excluding those 11, its accept share is 28\%; including them, 48\% (orals and posters 12 of 16, rejections and withdrawals 7 of 24, low-scored rejections 0 of 8), against 82\% on the ICLR 2027 main sample. Median percentiles on the ICLR 2026 sample and the ICLR 2027 main sample: Opus 4.6 35 and 55, Opus 5.5 30 and 45, GPT-6.1 Sol 45 and 56.5, GPT-6 Astra 49 and 58, Gemini 4 Argon 42 and 58. The two samples are stratified differently, by decision and by initial human score.

\paragraph{Version gap by paper set and posting month.} On the 10 decided ICLR 2026 papers that neither Opus endpoint recognized by probe or statement, both accept none. Within ICLR 2027, the Opus gap is $-50$, $-44$ and $-53$ points for the 4, 16 and 59 submissions first posted on arXiv before April 2026, in April to June, and from July on.

\paragraph{Stated recognition by decision class.} On the 39 decided ICLR 2026 papers with reviews from both Opus endpoints, accepted papers among those the model stated it recognized, against those it did not, by class (orals; posters; rejections with mean rating $\ge 5$; rejections with mean rating $\le 3$; withdrawals). Opus 4.6: 4/4 against 3/4; 3/3 against 2/5; 3/3 against 1/5; none stated against 0/8; 1/1 against 2/6. Opus 5.5: 4/5 against 3/3; 1/2 against 0/6; 0/4 against 0/4; 0/1 against 0/7; 0/2 against 0/5. The predicted counts in \cref{sec:boundaries} apply the not-stated rate of each class to the stated papers of that class. On the 18 papers flagged by the probes of Opus 5.5 and neither flagged nor claimed by Opus 4.6, Opus 5.5 and Opus 4.6 accepted 3/4 and 3/4 orals, 0/4 and 2/4 posters, 0/4 and 1/4 rejections with mean rating $\ge 5$, 0/2 and 0/2 rejections with mean rating $\le 3$, and 0/4 and 2/4 withdrawals.

\paragraph{arXiv posting date.} Within the ICLR 2027 submissions we reviewed, we regress each model's percentile on the mean human rating and compare residuals between submissions first posted on arXiv up to June 2026 and from July 2026 on. Later submissions receive higher residual percentiles from every model (Opus 4.6 $+4.5$, 79 papers; Opus 5.5 $+6.2$, 123; GPT-6.1 Sol $+6.1$, 143; GPT-6 Astra $+6.0$, 143; Gemini 3.8 Flash $+13.6$, 84; Gemini 4 Hydrogen $+7.5$, 84). The second registered plan tests this pattern on a fresh random sample (Q4).

\section{Robustness checks}
\label{app:robust}

\paragraph{Full text.} With references and appendices and without truncation, GPT-6 Astra gave the same rating as in the main run on all 22 papers that fit the client's input limit, GPT-6.1 Sol on 21, and Claude Opus 5.5 on all 24.

\paragraph{Reasoning effort.} Claude Opus 5.5 at medium effort gave the same rating as at xhigh on 21 of 24 papers (accept side 4 against 3 of 24), and the two runs' percentiles correlate at 0.86.

\paragraph{Percentile without a rating.} In a fresh session that shows only the paper and asks for its percentile and acceptance probability, without a review or rating, the percentiles correlate with the original ones at 0.78 (Opus 4.6, 22 papers), 0.90 (Opus 5.5), 0.52 (GPT-6.1 Sol) and 0.54 (GPT-6 Astra). Against the two-human reference of \cref{sec:ranking} they reach 0.17, 0.31, 0.37 and 0.31, against 0.23, 0.39, 0.17 and 0.43 for the original percentiles (paired differences $-0.05$, $-0.07$, $+0.20$ and $-0.12$, all 95\% intervals containing 0). Median percentiles without a rating are 65 (Opus 4.6), 41 (Opus 5.5), 62 (GPT-6.1 Sol) and 61.5 (GPT-6 Astra), against 55, 45, 56.5 and 58 after the rating.

\paragraph{Re-runs.} Re-running the endpoint models moved 1 of 22 (Opus 4.6), 1 of 24 (Opus 5.5), 5 of 24 (Gemini 3.8 Flash; 2 up, 3 down) and 3 of 24 (Gemini 4 Hydrogen; 1 up, 2 down) papers across the threshold; a full re-run of the two GPT models kept all 24 ratings.

\paragraph{Cross-scoring cells and pre-specified thresholds.} Accept share (\%) by critique source and scorer, written as earlier critique scored by the earlier/later version, then later critique scored by the earlier/later version. Critique only: Opus, 22 papers, 36/64 and 0/7; Opus, 57 papers, 25/44 and 0/7; Gemini, 24 papers, 56/54 and 2/6; GPT, 24 papers, 8/6 and 2/4. Paper and critique: Opus, 22 papers, 64/36 and 14/14; GPT, 24 papers, 38/17 and 8/4. Before the calls on the 57 Opus papers, the Gemini and GPT pairs, the paper-visible runs, the third-party scorer, the rewritten critiques and the fact check, we wrote down thresholds for the critique effect (at most $-25$ points: present; at least $-10$: absent) and the scorer effect (at most $-25$: later scorer harsher; at least $-10$: not harsher), with values in between undecided, and, for the paper-visible runs, for the share of the earlier version's verdict-critique gap recovered with the paper (at least 0.5: mainly the missing paper; at most 0.2: mainly the critique). Under these thresholds the critique effect is present on the 57 Opus papers and the Gemini pair and absent on the GPT pair without the paper, and in all three critique-only runs the scorer effect is above $-10$ points, the registered bound for a later scorer that is not harsher. With GPT-6.1 Sol as scorer of the 79 Opus critique pairs, the critique effect is present; after the critiques are rewritten into the neutral template it is $-24$ points ($[-34,-16]$), between the two thresholds and with the whole interval below $-10$. In the fact check of 30 papers, the share of weaknesses contradicted by the paper text is 2.8 points lower for Opus 5.5 than for Opus 4.6 ($[-6.4,+0.6]$), which meets the threshold for the later version being no less accurate (at most $+5$ points). With the paper, the recovered share is 0.60 for Opus 4.6 and 1.40 for GPT-6 Sol, the critique effect is present for Opus and between the thresholds for GPT ($-21$, $[-33,-8]$), and the scorer effect lies between the thresholds for both ($-14$ and $-12$).

\paragraph{Scorer dependence of the critique term.} The critique term in \cref{tab:cross} averages over the two scorers. Under the earlier and the later scorer it is, with the interaction (later minus earlier) and a 2{,}000-draw paper bootstrap interval: critique only, Opus, 22 papers, $-36$ and $-57$ ($-20$, $[-41,0]$); Opus, 57 papers, $-25$ and $-37$ ($-12$, $[-26,0]$); Gemini, 24 papers, $-54$ and $-48$ ($+6$, $[-8,+23]$); GPT, 24 papers, $-6$ and $-2$ ($+4$, $[0,+10]$); paper and critique, Opus, 22 papers, $-50$ and $-23$ ($+27$, $[+9,+45]$); GPT, 24 papers, $-29$ and $-12$ ($+17$, $[0,+38]$).

\paragraph{Exclusions and missing reviews.} Drift in accept share (points) under the rule used in the paper, which excludes probe hits and reviews that self-report recognition, and with self-reported recognitions kept; then the range when every missing or excluded review is set to accept or to reject for either version. Main sample: Opus $-68$ and $-67$, range $[-71,-62]$; Gemini $-71$ and $-71$. Second batch: Opus $-44$ and $-43$, range $[-43,-38]$; Gemini $-55$ and $-55$. Random sample (plan 3): Opus $-52$ and $-51$, range $[-52,-44]$, including three Opus 5.5 reviews blocked by the provider; Gemini $-53$ and $-53$. No probe hit occurred for these four models, and Gemini has no missing or self-reported reviews.

\paragraph{Re-anchoring and the human side.} On the second-batch pairs (Opus 56, Gemini 60 papers; human side: mean of three initial ratings $\ge 2.5$, 12 and 15 papers), balanced accuracy of the earlier/later version is 0.61/0.67 and 0.63/0.56 for raw ratings. Relative to raw ratings, per-version top-30\% cutoffs raise version agreement by 26 ($[+12,+38]$) and 38 points ($[+21,+50]$) and change balanced accuracy by $+0.02$/$+0.00$ and $-0.07$/$+0.03$; cutoffs fitted on the main sample raise agreement by 23 ($[+7,+39]$) and 32 points ($[+18,+45]$) and change balanced accuracy by $-0.13$/$-0.01$ and $-0.08$/$-0.02$. All balanced-accuracy intervals include zero (2{,}000-draw paired paper bootstrap).

\paragraph{Residual verdict phrases in stripped critiques.} Stripping removes sentences with scores, acceptance, rejection or recommendations. A regular expression for other overall-verdict phrases (for example ``merits publication'', ``prevent a higher rating'', ``ready for a top venue'') matches 9 of 79 stripped Opus 4.6 critiques, 2 of 79 Opus 5.5 critiques, and none of the 48 Gemini or 48 GPT critiques. Excluding every paper with a match, the critique effect is $-46$ points on 19 main-sample papers ($[-64,-26]$) and $-28$ on 49 second-batch papers ($[-38,-17]$), and the scorer effect $+12$ and $+13$.

\paragraph{Second labeler.} GPT-6 Astra relabeled a random sample of 400 weakness items with the same taxonomy and no access to the first labels. The labelers agree on 346 items (86.5\%; Cohen's $\kappa=0.85$, 95\% CI $[0.81, 0.89]$).

\paragraph{Sub-scores.} Across the reviews by the four primary models on all paper sets, every review with soundness 1 has a rating of 1 or 2; the GPT models assign soundness 1 in 22\% to 24\% of reviews against 7\% to 8\% for Claude Opus 5.5 and Gemini 4 Argon, and give a 2 in 87\% to 90\% of their reviews with soundness 2.

\section{Limitation-deletion edit}
\label{app:disclosure}
Candidates were the 27 public papers (15 award papers and 12 decided ICLR 2026 papers) that GPT-6.1 Sol, GPT-6 Astra and Claude Opus 5.5 had each reviewed without recognizing. GPT-6 Astra marked, by line number in the text the models see, the authors' statements of weaknesses of their own work and length-matched background passages. GPT-6.1 Sol then classified every marked limitation passage as a stated weakness, a main result or contribution, problem setup, or other, and only stated weaknesses were kept; the authors read the marked passages of four papers and removed four further passages from two of them. The 21 papers with at least 300 characters of limitation text form the sample. Background passages match the deleted length to within 30\%. Each version of each paper was reviewed twice by each GPT model and once by Claude Opus 5.5 (limitation and background deletions). Two re-runs of the same unedited paper differ by a median of 2 GPT percentile points; against the background deletion, two papers rose by 3 or more points and four fell by as much.

%% file: main.bbl
\begin{thebibliography}{51}
\providecommand{\natexlab}[1]{#1}
\providecommand{\url}[1]{\texttt{#1}}
\expandafter\ifx\csname urlstyle\endcsname\relax
  \providecommand{\doi}[1]{doi: #1}\else
  \providecommand{\doi}{doi: \begingroup \urlstyle{rm}\Url}\fi

\bibitem[Alharbi(2026)]{alharbi2026scrutinise}
Alharbi, E.
\newblock {Do large language models scrutinise what they review? A multimodal
  audit of scoring calibration, error detection, and author-identity effects}.
\newblock arXiv:2608.28626, 2026.

\bibitem[Baumann et~al.(2026)Baumann, Pei, Koyejo, and Hovy]{baumann2026stop}
Baumann, J., Pei, J., Koyejo, S., and Hovy, D.
\newblock {Stop Automating Peer Review Without Rigorous Evaluation}.
\newblock arXiv:2605.03202, 2026.

\bibitem[Beygelzimer et~al.(2023)Beygelzimer, Dauphin, Liang, and
  Vaughan]{beygelzimer2023neurips}
Beygelzimer, A., Dauphin, Y.~N., Liang, P., and Vaughan, J.~W.
\newblock {Has the Machine Learning Review Process Become More Arbitrary as the
  Field Has Grown? The NeurIPS 2021 Consistency Experiment}.
\newblock arXiv:2306.03262, 2023.

\bibitem[Biswas et~al.(2026)Biswas, Schoepp, Vasan, Opipari, Zhang, Hu, Joseph,
  Lease, Li, Stone, Wagstaff, Taylor, and Jenkins]{biswas2026aaai}
Biswas, J., Schoepp, S., Vasan, G., Opipari, A., Zhang, A., Hu, Z., Joseph, S.,
  Lease, M., Li, J.~J., Stone, P., Wagstaff, K.~L., Taylor, M.~E., and Jenkins,
  O.~C.
\newblock {AI-Assisted Peer Review at Scale: The AAAI-26 AI Review Pilot}.
\newblock arXiv:2604.13940, 2026.

\bibitem[Camelo-Guerrero \& Diaz-Rodriguez(2026)Camelo-Guerrero and
  Diaz-Rodriguez]{camelo2026closely}
Camelo-Guerrero, A. and Diaz-Rodriguez, J.
\newblock {How Closely Do LLM Reviews Align with Human Peer Review?}
\newblock arXiv:2608.03659, 2026.

\bibitem[Chauvin et~al.(2026)Chauvin, Merrer, Ta{\"\i}ani, and
  Tredan]{chauvin2026logprob}
Chauvin, T., Merrer, E.~L., Ta{\"\i}ani, F., and Tredan, G.
\newblock {Log Probability Tracking of LLM APIs}.
\newblock In \emph{International Conference on Learning Representations}, 2026.

\bibitem[Chen et~al.(2023)Chen, Zaharia, and Zou]{chen2023chatgpt}
Chen, L., Zaharia, M., and Zou, J.
\newblock {How is ChatGPT's behavior changing over time?}
\newblock arXiv:2307.09009, 2023.

\bibitem[Chen et~al.(2026)Chen, Yu, Liao, Jin, and Li]{chen2026calibrated}
Chen, Z., Yu, L., Liao, X., Jin, H., and Li, Q.
\newblock Position: Peer review should be calibrated via {LLM} scoring.
\newblock In \emph{Proceedings of the 43rd International Conference on Machine
  Learning}, volume 306 of \emph{Proceedings of Machine Learning Research},
  pp.\  169773--169801, 2026.

\bibitem[Garg et~al.(2025)Garg, Prasad, Singhal, Kirtani, Mandal, and
  Kumar]{garg2025revieweval}
Garg, M.~K., Prasad, T., Singhal, T., Kirtani, C., Mandal, M., and Kumar, D.
\newblock {ReviewEval: An Evaluation Framework for AI-Generated Reviews}.
\newblock arXiv:2502.11736, 2025.

\bibitem[Georgantas(2026)]{georgantas2026bottleneck}
Georgantas, C.
\newblock {Intelligence Is Not the Bottleneck: Validating an LLM First-Pass
  Manuscript Score Against Peer-Review Outcomes}.
\newblock arXiv:2606.15887, 2026.

\bibitem[Gloire \& Mitra(2026)Gloire and Mitra]{gloire2026doubleblind}
Gloire, B.~M. and Mitra, P.
\newblock {Large Language Models Threaten Double-blind Review}.
\newblock arXiv:2608.05157, 2026.

\bibitem[Golchin \& Surdeanu(2023)Golchin and Surdeanu]{golchin2023timetravel}
Golchin, S. and Surdeanu, M.
\newblock {Time Travel in LLMs: Tracing Data Contamination in Large Language
  Models}.
\newblock arXiv:2308.08493, 2023.

\bibitem[Han \& Kim(2026)Han and Kim]{han2026swapping}
Han, S. and Kim, M.
\newblock {Swapping the Judge Changes the Ranking: A Preregistered Study of
  Cross-Provider Replication in LLM-Judged System Comparisons}.
\newblock arXiv:2611.04090, 2026.

\bibitem[Huang(2026)]{huang2026lottery}
Huang, F.
\newblock {The Review Lottery: Benchmarking an Observational Estimator of
  Peer-Review Noise (ICLR 2017--2025)}.
\newblock arXiv:2610.06591, 2026.

\bibitem[Huang \& Zhou(2026)Huang and Zhou]{huang2026paradigms}
Huang, F. and Zhou, H.
\newblock {Comparative Evaluation of LLM-Based Review Paradigms for Academic
  Paper Quality Assessment}.
\newblock arXiv:2610.41391, 2026.

\bibitem[Jiang \& Ng(n.d.)Jiang and Ng]{jiang2025agentic}
Jiang, Y. and Ng, A.
\newblock {Agentic Reviewer: Technical Overview}.
\newblock \url{https://paperreview.ai/tech-overview}, accessed 14 November
  2026, n.d.

\bibitem[Jin et~al.(2024)Jin, Zhao, Wang, Chen, Zhu, Xiao, and
  Wang]{jin2024agentreview}
Jin, Y., Zhao, Q., Wang, Y., Chen, H., Zhu, K., Xiao, Y., and Wang, J.
\newblock {AgentReview: Exploring Peer Review Dynamics with LLM Agents}.
\newblock arXiv:2406.12708, 2024.

\bibitem[Karger et~al.(2025)Karger, Bastani, Yueh-Han, Jacobs, Halawi, Zhang,
  and Tetlock]{karger2025forecastbench}
Karger, E., Bastani, H., Yueh-Han, C., Jacobs, Z., Halawi, D., Zhang, F., and
  Tetlock, P.~E.
\newblock {ForecastBench: A Dynamic Benchmark of AI Forecasting Capabilities}.
\newblock In \emph{International Conference on Learning Representations}, 2025.

\bibitem[Keuper(2025)]{keuper2025injection}
Keuper, J.
\newblock {Prompt Injection Attacks on LLM Generated Reviews of Scientific
  Publications}.
\newblock arXiv:2509.10248, 2025.

\bibitem[Kim et~al.(2026)Kim, Deng, Vaughan, Su, Su, Agarwal, Li, Jaggi,
  Goldstein, Shah, and Dud{\'\i}k]{kim2026icml}
Kim, S. S.~Y., Deng, W.~H., Vaughan, J.~W., Su, B., Su, W., Agarwal, A., Li,
  S., Jaggi, M., Goldstein, D.~G., Shah, N.~B., and Dud{\'\i}k, M.
\newblock {Use and Effects of LLMs in Peer Review: A Randomized Experiment and
  Survey at ICML 2026}.
\newblock arXiv:2609.19420, 2026.

\bibitem[Konan et~al.(2026)Konan, Liu, and Liu]{konan2026paperscorer}
Konan, S., Liu, J., and Liu, Z.
\newblock {PaperScorer: How Predictable Is Paper Acceptance?}
\newblock arXiv:2610.31791, 2026.

\bibitem[Latona et~al.(2024)Latona, Ribeiro, Davidson, Veselovsky, and
  West]{latona2024lottery}
Latona, G.~R., Ribeiro, M.~H., Davidson, T.~R., Veselovsky, V., and West, R.
\newblock {The AI Review Lottery: Widespread AI-Assisted Peer Reviews Boost
  Paper Scores and Acceptance Rates}.
\newblock arXiv:2405.02150, 2024.

\bibitem[Li et~al.(2026)Li, Xiong, Zhu, Yu, Chen, Wang, Ma, and
  Lu]{li2026llmreviewer}
Li, L., Xiong, J., Zhu, C., Yu, R., Chen, C., Wang, J., Ma, R., and Lu, Z.
\newblock {LLM-as-a-Reviewer: Benchmarking Their Ability, Divergence, and
  Prompt Injection Resistance as Paper Reviewers}.
\newblock arXiv:2605.25415, 2026.

\bibitem[Li(2026)]{li2026whodrifted}
Li, Y.
\newblock {Who Drifted: the System or the Judge? Anytime-Valid Attribution in
  LLM Evaluation Pipelines}.
\newblock arXiv:2606.15474, 2026.

\bibitem[Liang et~al.(2023)Liang, Zhang, Cao, Wang, Ding, Yang, Vodrahalli, He,
  Smith, Yin, McFarland, and Zou]{liang2023feedback}
Liang, W., Zhang, Y., Cao, H., Wang, B., Ding, D., Yang, X., Vodrahalli, K.,
  He, S., Smith, D., Yin, Y., McFarland, D., and Zou, J.
\newblock {Can large language models provide useful feedback on research
  papers? A large-scale empirical analysis}.
\newblock arXiv:2310.01783, 2023.

\bibitem[Liang et~al.(2024)Liang, Izzo, Zhang, Lepp, Cao, Zhao, Chen, Ye, Liu,
  Huang, McFarland, and Zou]{liang2024monitoring}
Liang, W., Izzo, Z., Zhang, Y., Lepp, H., Cao, H., Zhao, X., Chen, L., Ye, H.,
  Liu, S., Huang, Z., McFarland, D.~A., and Zou, J.~Y.
\newblock {Monitoring AI-Modified Content at Scale: A Case Study on the Impact
  of ChatGPT on AI Conference Peer Reviews}.
\newblock In \emph{International Conference on Machine Learning}, 2024.

\bibitem[Liu \& Shah(2023)Liu and Shah]{liu2023reviewergpt}
Liu, R. and Shah, N.~B.
\newblock {ReviewerGPT? An Exploratory Study on Using Large Language Models for
  Paper Reviewing}.
\newblock arXiv:2306.00622, 2023.

\bibitem[Louis et~al.(2026)Louis, Vanleke, Ginis, and
  Algaba]{louis2026rebuttals}
Louis, M., Vanleke, T., Ginis, V., and Algaba, A.
\newblock {Rebuttals Move Peer-Review Scores, but Initial-Review Structure
  Bounds the Movement}.
\newblock arXiv:2606.22166, 2026.

\bibitem[Lu et~al.(2024)Lu, Lu, Lange, Foerster, Clune, and
  Ha]{lu2024aiscientist}
Lu, C., Lu, C., Lange, R.~T., Foerster, J., Clune, J., and Ha, D.
\newblock {The AI Scientist: Towards Fully Automated Open-Ended Scientific
  Discovery}.
\newblock arXiv:2408.06292, 2024.

\bibitem[Lu \& Yang(2026)Lu and Yang]{lu2026grade}
Lu, J. and Yang, S.
\newblock {The Grade Says Little, the Score Orders: AI Reviewer Panels Against
  a Venue's Ratings and Decisions}.
\newblock In \emph{NeurIPS 2026 Workshop on AI-Native Academia: Authorship,
  Peer Review, and Conference Governance under AI}, 2026.

\bibitem[Mao et~al.(2026)Mao, Zhang, Lun, and Yang]{mao2026dsar}
Mao, Z., Zhang, P., Lun, L., and Yang, H.
\newblock {When Critiques Do Not Change Decisions: DSAR for Threshold-Aligned
  LLM Reviewers}.
\newblock In \emph{Findings of the Association for Computational Linguistics:
  EMNLP 2026}, 2026.

\bibitem[Panickssery et~al.(2024)Panickssery, Bowman, and
  Feng]{panickssery2024recognize}
Panickssery, A., Bowman, S.~R., and Feng, S.
\newblock {LLM Evaluators Recognize and Favor Their Own Generations}.
\newblock In \emph{Advances in Neural Information Processing Systems}, 2024.

\bibitem[Preiss \& Yang(2026)Preiss and Yang]{preiss2026anchors}
Preiss, J. and Yang, Y.
\newblock {Guiding LLM Peer Reviewers: The Impact of Score Anchors on Review
  Evidence and Accuracy}.
\newblock arXiv:2609.01905, 2026.

\bibitem[Qin et~al.(2026)Qin, Jie, Qian, and Srifi]{qin2026interpreting}
Qin, W., Jie, R., Qian, Y., and Srifi, S.
\newblock {Interpreting LLM Scores in Peer Review: Alignment and Divergence
  Across Prompts, Domains, and Manuscript Features}.
\newblock arXiv:2610.35098, 2026.

\bibitem[Saha et~al.(2026)Saha, Juneja, Chaudhuri, Sajeevan, Shah, and
  Pruthi]{saha2026policies}
Saha, R., Juneja, G., Chaudhuri, D., Sajeevan, N., Shah, N.~B., and Pruthi, D.
\newblock {Policies Permitting LLM Use for Polishing Peer Reviews Are Currently
  Not Enforceable}.
\newblock In \emph{International Conference on Machine Learning}, 2026.

\bibitem[Sharma et~al.(2026)Sharma, Joachims, and Dean]{sharma2026favor}
Sharma, V., Joachims, T., and Dean, S.
\newblock {Do LLMs Favor LLMs? Quantifying Interaction Effects in Peer Review}.
\newblock arXiv:2601.20920, 2026.

\bibitem[Sunkavalli(2026)]{sunkavalli2026raters}
Sunkavalli, V.~K.
\newblock {LLM Judges as Raters: A Pre-Registered Audit of Severity, Halo,
  Reliability, and Version Instability in LLM Essay Scoring on Public Corpora}.
\newblock arXiv:2608.29517, 2026.

\bibitem[Thakkar et~al.(2025)Thakkar, Yuksekgonul, Silberg, Garg, Peng, Sha,
  Yu, Vondrick, and Zou]{thakkar2025feedback}
Thakkar, N., Yuksekgonul, M., Silberg, J., Garg, A., Peng, N., Sha, F., Yu, R.,
  Vondrick, C., and Zou, J.
\newblock {Can LLM Feedback Enhance Review Quality? A Randomized Study of 20K
  Reviews at ICLR 2025}.
\newblock arXiv:2504.09737, 2025.

\bibitem[Vasu et~al.(2026)Vasu, Joshi, Christmann, Sheth, Binkyte, Chowdhury,
  Kerkouche, Wang, and Fritz]{vasu2026style}
Vasu, S. S.~M., Joshi, S.~S., Christmann, P., Sheth, I., Binkyte, R.,
  Chowdhury, K.~D., Kerkouche, R., Wang, H.-P., and Fritz, M.
\newblock {Style over Substance? Linguistic Bias in LLM-based Peer Review}.
\newblock arXiv:2610.35254, 2026.

\bibitem[Yang et~al.(2026{\natexlab{a}})Yang, Hou, and
  Yang]{yang2026judgechanges}
Yang, Z., Hou, Y., and Yang, X.
\newblock {When the Judge Changes, So Does the Measurement: Auditing
  LLM-as-Judge Reliability}.
\newblock arXiv:2607.08535, 2026{\natexlab{a}}.

\bibitem[Yang et~al.(2026{\natexlab{b}})Yang, Li, Wang, Dobbie, and
  Liu]{yang2026believe}
Yang, Z., Li, H., Wang, Z., Dobbie, G., and Liu, J.
\newblock {Can We Believe AI Peer Review?}
\newblock arXiv:2610.19339, 2026{\natexlab{b}}.

\bibitem[Yang et~al.(2026{\natexlab{c}})Yang, Zhou, Wang, Jiang, and
  Zhang]{yang2026publications}
Yang, Z., Zhou, X., Wang, H., Jiang, Y., and Zhang, X.
\newblock Large language models in peer review: Decision alignment, review-text
  characteristics, and human--{AI} aggregation at {ICLR} 2025.
\newblock \emph{Publications}, 14\penalty0 (3):\penalty0 55,
  2026{\natexlab{c}}.
\newblock \doi{10.3390/publications14030055}.

\bibitem[Ye et~al.(2026)Ye, Cui, and Hadfield-Menell]{ye2026roleconfusion}
Ye, C., Cui, J., and Hadfield-Menell, D.
\newblock {Prompt Injection as Role Confusion}.
\newblock arXiv:2603.12277, 2026.

\bibitem[Ye et~al.(2025)Ye, Wang, Huang, Chen, Zhang, Moniz, Gao, Geyer, Huang,
  Chen, Chawla, and Zhang]{ye2025justice}
Ye, J., Wang, Y., Huang, Y., Chen, D., Zhang, Q., Moniz, N., Gao, T., Geyer,
  W., Huang, C., Chen, P.-Y., Chawla, N.~V., and Zhang, X.
\newblock {Justice or Prejudice? Quantifying Biases in LLM-as-a-Judge}.
\newblock In \emph{International Conference on Learning Representations}, 2025.

\bibitem[Ye et~al.(2024)Ye, Pang, Chai, Chen, Yin, Xiang, Dong, Shao, and
  Chen]{ye2024risks}
Ye, R., Pang, X., Chai, J., Chen, J., Yin, Z., Xiang, Z., Dong, X., Shao, J.,
  and Chen, S.
\newblock {Are We There Yet? Revealing the Risks of Utilizing Large Language
  Models in Scholarly Peer Review}.
\newblock arXiv:2412.01708, 2024.

\bibitem[Yu et~al.(2026)Yu, Zuccon, and Leelanupab]{yu2026backbone}
Yu, C., Zuccon, G., and Leelanupab, T.
\newblock {The Impact of Backbone Evolution on LLM-Based Relevance
  Assessments}.
\newblock arXiv:2610.09820, 2026.

\bibitem[Zhang \& Yang(2026)Zhang and Yang]{zhang2026versionshift}
Zhang, G. and Yang, H.
\newblock {LLM-as-a-Judge Version Shift: The Instrument Changed, Not the
  Standard}.
\newblock arXiv:2610.32986, 2026.

\bibitem[Zhang \& Zhang(2026)Zhang and Zhang]{zhang2026deanon}
Zhang, L. and Zhang, H.
\newblock {De-Anonymization at Scale via Tournament-Style Attribution}.
\newblock arXiv:2601.12407, 2026.

\bibitem[Zhang et~al.(2025)Zhang, Zhang, Ji, Hua, Haber, Cao, and
  Liang]{zhang2025pairwise}
Zhang, Y., Zhang, H., Ji, W., Hua, T., Haber, N., Cao, H., and Liang, W.
\newblock {From Replication to Redesign: Exploring Pairwise Comparisons for
  LLM-Based Peer Review}.
\newblock arXiv:2506.11343, 2025.

\bibitem[Zubeldia et~al.(2026)Zubeldia, Bolliet, Villaescusa-Navarro,
  Villanueva-Domingo, and Cranmer]{zubeldia2026auditing}
Zubeldia, {\'I}., Bolliet, B., Villaescusa-Navarro, F., Villanueva-Domingo, P.,
  and Cranmer, M.
\newblock {Auditing AI peer reviewers: a benchmark on real scientific papers}.
\newblock arXiv:2611.17052, 2026.

\bibitem[{\.Z}urawicki et~al.(2026){\.Z}urawicki, Farganus, Gawe{\l},
  Bystro{\'n}ski, and Kajdanowicz]{zurawicki2026praib}
{\.Z}urawicki, K., Farganus, J., Gawe{\l}, A., Bystro{\'n}ski, M., and
  Kajdanowicz, T.~J.
\newblock {PRAIB: Peer Review AI Benchmark of Behaviour of LLM-Assisted
  Reviewing}.
\newblock arXiv:2605.29815, 2026.

\end{thebibliography}
